\documentclass[letterpaper]{article} 
\usepackage{aaai2026}  
\usepackage{times}  
\usepackage{helvet}  
\usepackage{courier}  
\usepackage[hyphens]{url}  
\usepackage{graphicx} 
\usepackage{natbib}  
\usepackage{caption} 
\usepackage{algorithm}
\usepackage{algorithmic}

\usepackage{lscape}
\usepackage{tabularx,booktabs,array,ragged2e,caption}
\newcolumntype{L}[1]{>{\RaggedRight\arraybackslash}p{#1}}

\usepackage{tikz}
\usetikzlibrary{trees, positioning, shapes, arrows, shadows, decorations.pathmorphing, fit, backgrounds}

\usepackage{enumitem}
\usepackage{newfloat}
\usepackage{listings}
\DeclareCaptionStyle{ruled}{labelfont=normalfont,labelsep=colon,strut=off} 
\floatstyle{ruled}
\newfloat{listing}{tb}{lst}{}
\floatname{listing}{Listing}
\nocopyright 

\title{Limits of Safe AI Deployment: Differentiating Oversight and Control}

\author{David Manheim \textsuperscript{\rm 1,2}, Aidan Homewood \textsuperscript{\rm 3}} 
\affiliations{
     \textsuperscript{\rm 1}Association for Long Term Existence and Resilience\\
    \textsuperscript{\rm 2} Technion - Israel Institute of Technology\\
    \textsuperscript{\rm 3} Centre for the Governance of AI, London, UK
}
 \date{\today}

\begin{document}

\maketitle

\begin{abstract}
Oversight and control, which we collectively call supervision, are often discussed as ways to ensure that AI systems are accountable, reliable, and able to fulfill governance and management requirements. However, the requirements for "human oversight" risk codifying vague or inconsistent interpretations of key concepts like oversight and control. This ambiguous terminology could undermine efforts to design or evaluate systems that must operate under meaningful human supervision. This matters because the term is used by regulatory texts, such as the EU AI Act, the NIST AI Risk Management Framework, and the OECD AI Principles.

This paper undertakes a targeted critical review of literature on supervision outside of AI, along with a brief summary of past work on the topic related to AI. We then differentiate control as being ex-ante or real-time, and operational rather than policy or governance. In contrast, we view oversight as either performed ex-post, or a policy and governance function. We suggest that control aims to prevent failures, while oversight focuses on detection, remediation, or incentives for future prevention, and we note that preventative oversight strategies necessitate control.

Building on this foundation, we make three contributions. First, we  propose a framework to align regulatory expectations with what is technically and organizationally plausible, articulating the conditions under which each mechanism is possible, where they fall short, and what is required to make them meaningful in practice. Second, we outline how supervision methods should be documented and integrated into risk management, and drawing on the Microsoft Responsible AI Maturity Model, we outline a maturity model for AI supervision. Third, we explicitly highlight some boundaries of these mechanisms, including where they apply, where they fail, and where it is clear that no existing methods suffice. This foregrounds the question of whether meaningful supervision is possible in a given deployment context, and can support regulators, auditors, and practitioners in identifying both present and future limitations. 
\end{abstract}

\section{Background}

Oversight and control are frequently invoked as mechanisms for ensuring AI system safety, including in regulation, yet the terms are conflated or used interchangeably, in part because their meaning is often unclear, inconsistent, or disputed. Clarifying this is critical for legal requirements and standards which require either, or both. While this is a seemingly narrow component of governance, reliance on oversight and control as a central part of governance is nearly universal, including in American policy \cite{whitehouse2023eo} and risk management frameworks \cite{nist2023rmf}, in the EU AI Act, \cite{eu_ai_act_2024}, in OECD principles for AI, \cite {oecd2024AIPrinciples} and in various international standards. \cite{isoiec22989_2022,isoiec23894_2023}

We begin by considering the terms and what they mean based on a review of both management literature and AI-specific discussions of the terms. The review leads to a conclusion that control refers to operational management and technical methods taken before or during deployment to prevent failures. In contrast, oversight can be passive, post-hoc operational processes or governance methods. It can also be active, that is, methods which enable human control to prevent failure. 

This clarity about meaning allows us to provide further explanation the processes which are used to accomplish the goals of supervision of AI systems, and how they must be reported. Finally, we provide a maturity model to indicate the extent to which the goals of supervision of AI systems are accomplished. Hopefully, these three pieces - clarity about what is required, reporting of what is done, and evaluation of success - provide the necessary conditions needed for clear standards, effective regulation, and eventually, safe AI systems.

\section{Literature Review}

To start, we note that supervision is a governance requirement that applies in many domains, and is only one of several necessary approaches, as illustrated in figure \ref{fig:high_level_typology}. Nonetheless, AI Governance has specific issues, so our review seeks to understand the ideas first in the context of management and governance generally, then within the specific context of AI system supervision.

\begin{figure}[htbp]
\centering
\scalebox{0.6}{
\begin{tikzpicture}[
    system/.style={rectangle, rounded corners=8pt, draw=teal!80, fill=teal!20, very thick, text centered, inner sep=10pt, drop shadow},
    sys_f1/.style={rectangle, rounded corners=8pt, draw=teal!40, fill=teal!10, very thick, text centered, text=gray, draw opacity=0.5, inner sep=10pt},
    sys_f2/.style={rectangle, rounded corners=8pt, minimum width=60pt, minimum height=25pt, draw=teal!20, fill=teal!5, very thick, text centered, draw opacity=0.1, inner sep=10pt},
 control/.style={rectangle, rounded corners=8pt, draw=blue!80, fill=blue!20, thick, text centered, inner sep=8pt, drop shadow},
    oversight/.style={rectangle, rounded corners=8pt, draw=green!80, fill=green!20, thick, text centered, inner sep=8pt, drop shadow},
    main_edge/.style={->, thick, draw=gray!50},
    node distance=1.8cm,
    level 1/.style={sibling distance=6cm}, 
    level 2/.style={sibling distance=4cm}
]

\node (origin) [system] {Governance};

\node (supervision) [system, below=of origin] {Supervision};

\node (principles) [sys_f1,draw opacity=0.5, below left=of origin] {Principles};

\node (culture) [sys_f1,draw opacity=0.5, below right=of origin] {Culture};



\node (control) [control, below=of supervision, xshift=-4cm] {Control};
\node (oversight) [oversight, below=of supervision, xshift=3cm] {Oversight};

\draw [main_edge] (origin) -- (supervision);
\draw [main_edge] (origin) -- (principles);
\draw [main_edge] (origin) -- (culture);
\draw [main_edge] (supervision) -- (control);
\draw [main_edge] (supervision) -- (oversight);
\end{tikzpicture}
}
\caption{}
    \label{fig:high_level_typology} High-level typology of governance focused on control and oversight, which we collectively call supervision.
\end{figure}

In management, \citet{CullenBrennan2017} differentiate between control, monitoring, and oversight for boards of directors controlling companies - and, incidentally, show that boards cannot accomplish all of three of these goals. They use a grounded-theory approach to review the issue and find widespread terminological confusion, so that  “Some definitions use the three terms interchangeably,” and that “interpretation of the terms… varies”. To resolve this confusion, they say that control is “the capacity to initiate, constrain, circumscribe, or terminate action,” quoting \citet{Herman1981}. In contrast, they suggest that “Oversight means overseeing or monitoring others’ activities and not doing things directly,” quoting \citet{Kanda2000}. Their conclusion is that control involves both observation and the capacity for direct action, while oversight relies on indirect observation, with or without subsequent intervention.

In the AI context, \citet{TsamadosEtAl2024} discusses supervisory control versus teaming as forms of “operational human control” during deployment, using \citet{VerdiesenEtAl2021}’s framework for autonomous weapons systems, as shown in Figure \ref{fig:framework}. This differentiates timing of control (before, during, or after deployment) and governance. The time dimension has an obvious meaning for weapons systems, where the deployment is a fixed, presumably short time window, but the corresponding time frames for any AI system can be understood similarly. That is, before involves design decisions, during involves operational decisions, and after involves review. In the last case, even if the system is still running, post-deployment would involve looking at the system's past actions, rather than control and changing the outputs. However, \citet{TsamadosEtAl2024} focuses on timing to the exclusion of \citet{VerdiesenEtAl2021}’s critical dimension distinguishing governance, sociotechnical, and engineering layers of control—central for differentiating methods of oversight and control. 

\citet{TsamadosEtAl2024} also review (and agree with) a variety of criticisms of levels of autonomy as a way to understand human control. The criticisms include loss of situational awareness, contextual changes, and trust. The critiques can largely be explained because levels of autonomy almost exclusively adopts the Observe-Orient-Decide-Act (OODA) view of processes, where there is a single loop that humans are inside or outside of. Critically, however, levels of autonomy models misinterpret Boyd’s OODA loop by neglecting the complexity of the “orient” phase - which, in Boyd’s original work contains almost all of the complexity, as noted by \citet{Richards2011}. As a result, the levels of autonomy concept highlights issues with the decisions and actions portions of the loop, while the noted criticisms and failures of the levels of autonomy concept largely relate to failures of orientation. For this reason, \citet{TsamadosEtAl2024}’s focus on control means that it fails to include any implications for oversight, given that real-time supervisory control is often infeasible, as will be discussed later.

\citet{Enqvist2023} starts her review of the EU AI Act requirements for oversight by pointing out that “Human oversight has been much stressed and discussed as a safeguarding measure to ensure human centrism in AI deployment.” Similarly, \citet{MethnaniEtAl2021} notes that “‘Human oversight’ is one of the requirements being put forward as a means to support human autonomy and agency,” and proposes {\textit{variable autonomy}} as a way to ensure “meaningful human control.” Unfortunately, in both cases control and oversight are not differentiated or clearly defined. \citet{HoDacMartinez2024} go further by implicitly framing control as direct human action, while oversight denotes a broader supervisory role. In doing so, they start by asserting that “The adoption of human oversight measures makes it possible to regulate, to varying degrees and in different ways, the decision-making process of Artificial Intelligence (AI) systems.” The claim, which is present in the EU AI Act, is that oversight is effective in enabling regulation of AI. However, this is asserted without evidence, and without clearly defining the requirements. (The requirements are evidently left for the Code of Practice, which unfortunately also does not do so.) This means that standards groups like ISO and CEN/CENELEC are left to define (meaningful) oversight in ways that enable the regulation’s stipulated ability to provide meaningful reassurance. This is worrying, as noted by \citet{Koulu2020} among others, which mentions “human oversight in EU policy for controlling algorithmic systems” and notes that human-in-the-loop risks becoming an “empty procedural shell” that fails to ensure meaningful human oversight.

Finally, \citet{Bengio2024}, in his US Senate testimony, focused on catastrophic outcomes and underscored four areas where human choices shape AI risk; "(1) access—who can tinker with powerful AIs, what protocols must they follow, under what kind of oversight? (2) misalignment—the extent to which powerful AIs act not as intended by their developers, potentially causing severe or even catastrophic harm” as well as “(3) raw intellectual power… and (4) scope of actions—the ability to affect the world and cause harm in spite of society’s defenses.“  However, once again the technical question of how oversight is accomplished and whether it is effective is left unanswered.

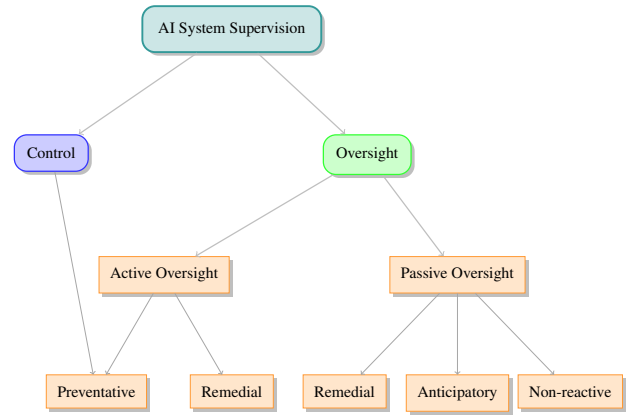
\begin{figure}[!htbp]
\centering
\scalebox{0.6}{
\begin{tikzpicture}[
    system/.style={rectangle, rounded corners=8pt, draw=teal!80, fill=teal!20, very thick, text centered, inner sep=10pt, drop shadow},
 control/.style={rectangle, rounded corners=8pt, draw=blue!80, fill=blue!20, thick, text centered, inner sep=8pt, drop shadow},
    oversight/.style={rectangle, rounded corners=8pt, draw=green!80, fill=green!20, thick, text centered, inner sep=8pt, drop shadow},
    oversight_dim/.style={diamond, draw=purple!80, fill=purple!20, thin, text centered, inner sep=5pt, aspect=2, drop shadow},
    oversight_type/.style={rectangle, draw=orange!80, fill=orange!20, thin, text centered, inner sep=7pt, drop shadow},
    example/.style={rectangle, dashed, draw=gray!60, fill=gray!15, thin, text centered, font=\small, text width=3cm, inner sep=5pt},
    main_edge/.style={->, thick, draw=gray!50},
    sub_edge/.style={->, thin, draw=gray!70},
    example_edge/.style={->, dotted, draw=gray!50},
    node distance=1.8cm,
    level 1/.style={sibling distance=6cm}, 
    level 2/.style={sibling distance=4cm}, 
    level 3/.style={sibling distance=3.5cm}, 
    level 4/.style={sibling distance=3cm}
]

\node (origin) [system] {AI System Supervision};

\node (control) [control, below=of origin, xshift=-4cm] {Control};
\node (oversight) [oversight, below=of origin, xshift=3cm] {Oversight};

\draw [main_edge] (origin) -- (control);
\draw [main_edge] (origin) -- (oversight);


\node (active) [oversight_type, below=of oversight, xshift=-4.5cm] {Active Oversight};
\node (passive) [oversight_type, below=of oversight, xshift=2cm] {Passive Oversight};

\draw [main_edge] (oversight) -- (active);
\draw [main_edge] (oversight) -- (passive);

\node (preventative_active) [oversight_type, below=of active, xshift=-1.5cm] {Preventative};

\node (remedial_active) [oversight_type, below=of active, xshift=1.5cm] {Remedial};

\draw [sub_edge] (control) -- (preventative_active);
\draw [sub_edge] (active) -- (preventative_active);
\draw [sub_edge] (active) -- (remedial_active);

\node (remedial_passive) [oversight_type, below=of passive, xshift=-2.5cm] {Remedial};
\node (nonreactive_passive) [oversight_type, below=of passive, xshift=2.5cm] {Non-reactive};
\node (anticipatory_passive) [oversight_type, below=of passive] {Anticipatory};

\draw [sub_edge] (passive) -- (anticipatory_passive);
\draw [sub_edge] (passive) -- (remedial_passive);
\draw [sub_edge] (passive) -- (nonreactive_passive);

\end{tikzpicture}
}
\caption{\label{fig:typology} Detailed typology of AI supervision methods}
\end{figure}

With that background, and keeping in mind the problem of proposing new ideas in already saturated fields \cite{Munroe2011}, we introduce domain-specific definitions of each of the two terms, and provide a typology which is outlined in Figure \ref{fig:typology}. We group them together as supervision - not to suggest that this additional term is useful on its own, but as a collective term for both methods for the below discussion.

\section{Defining Control}
\begin{figure}[htbp]
    \centering
    \includegraphics[width=0.50\textwidth]{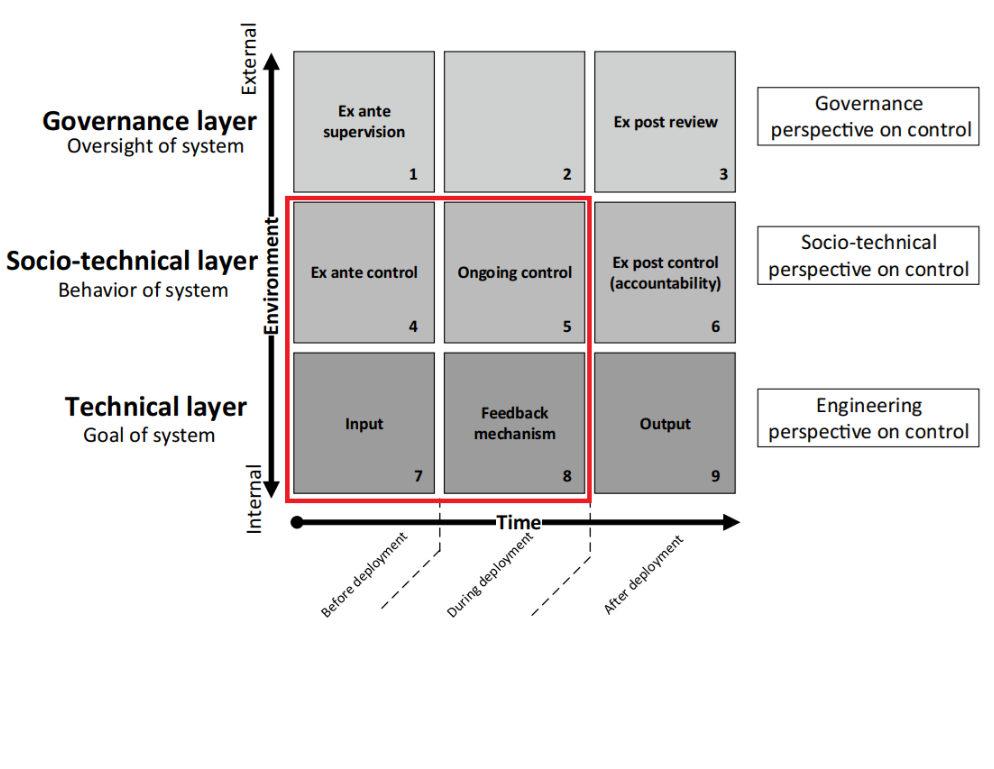} 
    \caption{ \label{fig:framework} The \citet{VerdiesenEtAl2021} framework with control-relevant portions outlined in red.}
\end{figure}

Using the above frameworks and varied definitions, control can only refer to actions taken before or during deployment, and to management and technical methods, as shown in Figure \ref{fig:framework} in red.

The simple cases are those discussed in the diagram directly, and we group technical and sociotechnical as operational mechanisms, rather than policy or governance mechanisms. Similarly, we exclude governance-level mechanisms from control. This means that control follows the contours of how the term is used in control theory. \cite{ogata2010modern} This partly includes work on what has recently been styled "AI Control" \cite{greenblatt2024ai} but our definition includes more prosaic risks of control failure and simpler systems, and excludes red-teaming and similar approaches proposed there, which we classify as oversight.

Note that there are also ways to build systems in which the output is controlled by some external constraint which function as control. Such external control can be an ex-post control measure - though as an operational function, it cannot be a governance intervention. However, it is better understood within the broader system that includes that external constraint as an ex-ante strategy implemented at either the technical or sociotechnical level. Similarly, we argue that what control theorists would call post-deployment feedback mechanisms are either put in place prior to deployment, or are later modifications triggered by a governance or oversight decision.

\section{Defining Oversight}
Oversight, which as we have said, exists either at a (non-operational) management or governance level, or is ex-post, or both, can be split into several classes; active versus passive, and preventative versus remedial, versus non-reactive.

An example of passive preventive oversight is anomaly detection, where humans are not involved on an ongoing basis, but are alerted and can intervene with a control once the anomaly is flagged. An example of preventative active oversight is red-teaming of a system before deployment, which can identify issues which are then presumably either remedied via a control strategy, or monitored via ex-post oversight. As both of these illustrate, preventative oversight ultimately requires control. 

On the other hand, there are a variety of remedial or passive oversight strategies. For example, post-hoc review and patching of incorrect decisions, liability for harms imposed, or the ability to appeal automated judgments to human review are each remedial, while accident and near-miss reporting are post-hoc. 

\section{ Control Strategies: Prerequisites and Limits}
Because control requires either real-time human monitoring or prior definition of what is and is not allowed, control depends on either clearly defined limitations, or immediate real-time human judgment about what is allowed. That is, ex-ante strategies require either explicit or implicit defined limits. In the latter case, to the extent that ex-ante limits and guarantees are incompletely defined or formalized, some ongoing oversight, often called “human in the loop” is necessary - but this has its own limitations, discussed below. In narrow application domains, there is often a simple expedient of strictly defining limits for the automated system, and enforcing these programmatically. For the control to be sufficient without human oversight, the control system requires a formalization of all relevant systems and possible scenarios where the limits need to apply - as occurs routinely in control system design. For example, low level automation of fuel injector pumps and similar simple systems which operate too quickly for human oversight are designed and controlled in this fashion. 

However, early basic work on control theory explains that controllability relies on a known system and sufficient feedback. \cite{kalman1960contributions, lin1974structural} If context changes or the scope of action is not constrained, designing controls for limiting failures, (much less controlling full autonomy,) will be impossible - though there is work towards “Safeguarded AI” which would enable reliable world models with provable claims, in place of ad-hoc application-specific limitations (\citet{DalrympleEtAl2024}). At least until that occurs, given the risks involved, \citet{MitchellEtAl2025} recently argued that fully autonomous general systems should simply never be built.

Lastly, we note that control for frontier AI systems is a developing field with many open questions, \cite{UKAISI2025ResearchAgenda} but it is still far from clear whether it can ever be done robustly. Some have argued it is fundamentally impossible, \cite{Brcic_2023} and many experts have suggested that it may require fundamental advances or changes in the design and architecture of systems. \cite{Bengio2025InternationalAI}

\subsection{Limits of Human in the Loop}
Human in the Loop (HITL) is a term often used for human control. This sometimes involves a human providing a predetermined input in an otherwise automated control loop, for example, a human whose job is to approve a predetermined computational decision. Even if situations in which the human should intervene or act differently are laid out beforehand, this provides a veneer of human oversight into what is essentially an automated process. \cite{Crootof2023Humans, Dekker2005} That is, in certain automated systems, humans are included to check a box, but the system is not actually built to enable or respect their autonomous judgment.

The issues with HITL control systems can make them less useful or even harmful, depending on the design and context - and this is not new, or unique to supervision of AI. \cite{bainbridge1983ironies} 
There can be well-designed systems that use HITL for effective control, but doing so requires intentional design, and a substantive argument that the system as constructed accomplishes this. 
Suggestions like "variable autonomy" from \citet{MethnaniEtAl2021} are potentially a useful path to this, at least in cases where the human can understand and meaningfully exercise judgment quickly enough. However, this approach blurs the distinction or bridges the gap between control and oversight. That is, variable autonomy HITL ranges from where the human is effectively functioning as passive oversight that actively intervenes when needed, to where the human always makes substantive decisions in the process, so that the AI system functions as decision support rather than automation. 

\section{Oversight Strategies: Prerequisites and Limits}
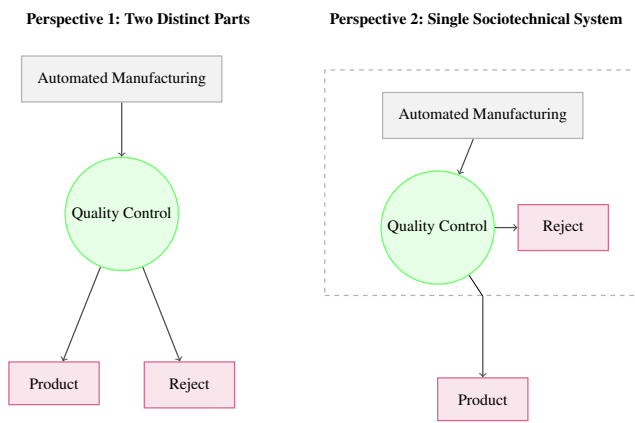
\begin{figure}[htbp]
\centering
\scalebox{0.6}{
\begin{tikzpicture}[
    component/.style={rectangle, draw=gray!60, fill=gray!10, thick, text centered, inner sep=10pt, minimum width=3cm}, 
    human/.style={circle, draw=green!60, fill=green!10, thick, text centered}, 
    system_boundary/.style={rectangle, draw=gray!60, dashed, thick, text centered, inner sep=25pt, minimum width=7cm, minimum height=5cm}, 
    process/.style={rectangle, draw=purple!60, fill=purple!10, thick, text centered, inner sep=10pt, minimum width=2cm}, 
    data_flow/.style={->, thick, draw=black!70},
    control_flow/.style={->, thick, draw=red!70, dashed},
    node distance=2cm and 4cm 
]

\begin{scope}[name=Dg1]
    \node [component] (automated_line) {Automated Manufacturing};
    \node [human, below=of automated_line, yshift=.8cm] (humanqc) {Quality Control};
    \node [process, below=of humanqc, xshift=-1.5cm] (product) {Product};
    \node [process, below=of humanqc, xshift=1.5cm] (bad_product) {Reject};

    \draw [data_flow] (automated_line) --(humanqc);
    \draw [data_flow] (humanqc) -- (product); 
    \draw [data_flow] (humanqc) -- (bad_product); 

    \node [above=0.5cm of automated_line.north west, anchor=south west, font=\bfseries] {Perspective 1: Two Distinct Parts};
\end{scope}

\begin{scope}[xshift=8cm, yshift =-2.3cm] 

    \node [system_boundary] (sociotechnical_system) {};
    \node [component, below=of sociotechnical_system.north, yshift=1.5cm] (automated_elements) {Automated Manufacturing}; 
    \node [human, above=of sociotechnical_system.south, anchor=south, yshift=-1.75cm, xshift=-1cm] (human_in_loop) {Quality Control}; 
    \node [process, below=of sociotechnical_system.south, yshift=.2cm] (sociotech_product) {Product};
    \node [process, right=of human_in_loop, xshift=-3.5cm] (sociotech_bad_product) {Reject};

    \draw [data_flow] (automated_elements) -- (human_in_loop);
    \draw [data_flow] (human_in_loop) -- (sociotech_bad_product);
    \draw (human_in_loop) -- (sociotechnical_system.south);

    \draw [data_flow](sociotechnical_system) -- (sociotech_product);

    \node [above=0.8cm of sociotechnical_system.north west, anchor=south west, font=\bfseries] {Perspective 2: Single Sociotechnical System};
\end{scope}
\end{tikzpicture}
} 
\caption{Comparing two possible structures for defining identical systems}
 \label{fig:comparison}
\end{figure}

The different varieties of oversight have different limitations. To discuss them, however, we must first be clear about system boundaries and the role they play in defining control and oversight. A simple example which could involve simple automation or AI system involvement is shown in Figure \ref{fig:comparison}. Here, an automated manufacturing line with human quality control can either be portrayed as two distinct parts, one that is automated, and one that is providing active oversight and remedial intervention, or portrayed as a single sociotechnical system with a human in the loop. 

How to categorize such a system is a decision on the part of management about how to document the supervision and governance of the system. This highlights that some distinctions drawn in the paper are themselves management and governance decisions, but they are not purely notational or semantic, as they can have implications for how the system is managed in practice. Regardless of classification, however, the method of supervision of the system should be clearly stated, and both the goals of supervision and the way in which the system accomplishes those goals should be clearly understood.

This brings us back to differentiating between control and oversight. We argue that if the output is controlled by some external system, the control should be evaluated in regards to the entirety of both the system and the “external” control system, and the sociotechnical system has a control component. Alternatively, we can say that the control system is a post-hoc remedial intervention, making it oversight of the production process, rather than control.

For example, we see that LLMs are capable of entertaining the idea of unacceptable outputs, but newer reasoning models have the LLM chain-of-thought output which filters this out, seemingly ex-post. Of course, this categorization is arguable, but if we view the reasoning model as a single system, the failure of control via chain of thought emphasizing compliance is a failure, rather than a lack of a control mechanism. The relevant system is therefore the full reasoning model, not the underlying LLM - and the chain-of-thought is a deployment-time feedback mechanism for outputs, not a post-deployment (non-human, automated) oversight system. Similarly, an LLM model might be capable of providing unacceptable output, but will not do so due to the system prompt, and a model that has undergone RLHF might be “capable” of certain classes of problematic output, but not actually provide those outputs due to the RLHF training. In each of the above cases, the control is exercised either before or during deployment, not afterwards. These are each ex-ante control strategies, and could therefore be interpreted as either technical or socio-technical control. In these cases, if a user accesses a model via an API and either, in the first scenario, uses a modified prompt that does not stop these outputs, or in the second scenario, fine-tunes the model to reverse the safety mechanisms, this is a failure or circumvention of the control, which is distinct from not having a control mechanism. This illustrates that ex-ante control can be harder to ensure than having a human actively reviewing and approving output, but can also be the only viable strategy due to speed of the system or the economics of requiring human involvement.

\subsection{Importance of Understanding for Oversight}
We must also discuss human understanding of the system, which is often taken for granted. In the general case of oversight, for example, as required by ISO 37000 for oversight of organizations, the person providing oversight must have sufficient understanding for the oversight to be meaningful. For organizations, oversight includes ensuring that internal control systems are appropriately designed and working. As a consequence, for oversight of automated and AI systems used in an organization, the same is true. 

The needed understanding generally includes the inputs and outputs of the system, the design and components of the system and the interactions between the subsystems, and how the system dynamically responds to different classes of change. Often, there are explicit sub-systems or monitoring devices to enable oversight. A simple example in an automobile is the speedometer, which provides a more accurate gauge of speed than human perception alone. For AI systems, a number of critical enablers exist, including AI explainability and AI interpretability, and tools that can enable these are often critical. Lastly, operational and design transparency are usually needed for understanding the AI systems operations. 

As the examples show, the details of what must be understood will differ greatly depending on system architecture and integration, and this deserves more attention. Thankfully, a vast literature on understanding and controlling systems (which we will not review here) exists, ranging from control theory for automated control, to work in systems engineering and management.  We review then in Section 3.2, and they are critical, but not explored in depth.

Understanding also helps provide the fundamental requirement of oversight, which is the ability to detect failures. Putting a human in the loop for an autonomous weapons system does not prevent mistaken usage \cite{Atherton2022Errors}, which is a failure of the oversight system, with critical legal implications. \cite{Jain2023Autonomous} Similarly, auditing logs for past errors is only effective if the auditing procedure can successfully identify what has happened. Even human review of events is far from perfect at finding if and when mistakes were made - either because they lack understanding needed to find a given type of failure, or because they are fallible. And the latter can be ameliorated by further review, while the former cannot.

Lastly, as noted by \citet{Goldenfein2024} oversight for governance or legal purposes sometimes requires not just human involvement, but human accountability. This goes beyond the scope of oversight and control itself, but is a requirement that can be enabled by oversight. That said, the requirements and limits laid out here do not necessarily provide assurance of accountability, just of meaningful oversight.

With those clarifications, we can discuss the additional limitations and requirements for oversight - beyond understanding and detection - considering first active, then passive methods. 

\subsection{Limits of Active Oversight}
Active oversight involves real-time human involvement in monitoring the system, and either intervening, or fixing identified failures. The former uses human oversight as a way to trigger direct or indirect human control of the system, for example, to take certain decisions out of the automated system's purview. (The latter is sometimes referred to as human-on-the-loop.) This is possible when the human oversight supervisor both understands the system and outputs, and can meaningfully intervene before problems due to automated decisions manifest. The first issue, of understanding, will be discussed below, but the second is primarily about the relative speed of the system and the negative impacts that must be prevented, and human attention and reaction time.

Some systems operate too rapidly for meaningful active oversight. An example might be low-level control of an automobile's fuel injection system. Others operate in ways that are not understood well enough to ensure active oversight is practical. For example, LLMs producing code changes in complex systems can be tested via unit tests, or formally guaranteed, but real-time oversight of the changes is not meaningful, since the impacts of code changes are very often not understood even by those actively developing the system, much less in cases where code is being modified for systems that are no longer being actively developed.

\subsection{Limits of Human on the Loop}
In complex systems, there are often a combination of strictly defined limits for fully automated parts of a process and real-time oversight and intervention as a control. For example, in aviation, automatic flight guidance systems have  specified altitude, speed, and weather conditions under which they may be operated. \citet{FAA2014AC25.1329} A similar situation exists for medical infusion pumps with dose ranges and flow rates, \cite{HoffmanBacon2020} or industrial robotics with safety cages \cite{KrugerEtAl2009}. In all three cases there is continuous human monitoring, and human error, miscommunication, or alert fatigue can still lead to failures.  \cite{Crootof2023Humans} Again, the ability to actually identify errors is the single critical enabler.

We reject Dekker's view that situational awareness is a post-hoc construction, and the contention, later partially walked back, that errors don't truly exist. \cite{Dekker2005} As noted earlier, we reiterate that the understanding of systems is a fundamental part of what is required for oversight, and believe that the "emic" perspective which \citet{Dekker2005} adopts from anthropology is useful for understanding, but does not replace the "etic" perspective. Instead, failures are when governance and/or ethical standards are violated, or when real world harms are inflicted, not post-hoc constructions. At the same time, we strongly endorse the points made about complacency being used as an excuse in place of acknowledging the conflicting safety and operational goals, and the rejection of automating our way out of needing oversight. Stated more plainly by \citet{Crootof2023Humans}, "humans aren’t a regulatory or design patch to be haphazardly inserted as a solution to problems that are really about the way the system as a whole is structured." 

\subsection{Limits of Passive Oversight}
In many cases, systems are constructed to either have sufficient active oversight to control real-time risks, or to not need such oversight. In some cases, this is because errors can be retrospectively fixed, at an acceptably low cost. For example, banking fraud is sometimes caught in real time, but in other cases occurs and can be caught in time for banks to then reverse the transaction. However, in the majority of cases of passive oversight, it serves a very different purpose, and should not be conflated with either control, or active oversight, each of which can stop failures, or remedial passive oversight, which corrects them. Instead, as noted earlier, passive oversight is often anticipatory, and can help minimize risks ex-ante via incentives, or is intended to provide information which can reduce preventable future failures.

As noted above, in all oversight, the sine qua non is the ability to detect failures. Given that, the primary limitation of retrospective passive oversight is the ability to actually reverse the action, or at least remediate and mitigate the consequences sufficiently. Critical factors also include latency and the speed at which harms are fixed. Critically, it is sometimes the case that feedback masks errors, as seen in the case of predictive policing, \cite{ensign2018runaway} and even when errors can be found, the detection may be long-delayed. \cite{Akpinar2021Differential}

For non-reactive and anticipatory passive oversight, the goal is to identify failures after they occur. Despite not preventing the failure, post-hoc strategies can still be valuable as ways to (1) incentify management to minimize risks ex-ante; (2) enable failures not to be repeated; and (3) serve as warnings to other firms in ways that reduce the incidence or severity of future events. It is also worth noting that the way in which oversight without control operates in each of these non-preventative oversight cases provides a strong public-policy argument against allowing undisclosed accidents or no-fault settlements, when such strategies are used.

The methods also have significant issues with honest disclosures and firm's incentives to hide or at least not aggressively look for failures. For this reason, the safety culture of firms is a critical enabler \cite{Manheim2023SafetyCulture}, and external review and audit standards may be necessary for firms to declare that these forms of {\it meaningful} passive oversight are in place. \cite{Manheim2025Necessity} 

Lastly, these passive methods may discharge management or governance responsibilities, but are insufficient for preventing loss of control of AI systems, which is especially concerning when they are deployed in safety critical systems where accidents can cause loss of life, or in cases where more general AI systems have the potential to create larger scale disasters. \cite{russell2019human, cave2018ai, dafoe2023open} 

\section{Contexts Where Oversight and Control May Be Unnecessary}
In the discussion above, it is possible to mistakenly assume that oversight and control must always either be sufficient, or insufficient. When this is untrue, it should be clearly stated, and the reasons why explicated, including the costs and benefits involved. "Balancing risk and reward" is a well-worn and often useless adage, and should be deployed carefully - not as an excuse, but as part of actual analysis. As \citet{NIST2023} notes in the context of trustworthiness and the decision to deploy AI systems, this should be "a contextual assessment of... the relative risks, impacts, costs, and benefits..." informed by a broad set of interested parties." In many cases, however, a complex analysis and/or mitigations are unnecessary, or infeasible at the likely cost, and this should be stated clearly.

For example, open-weight generative AI chatbots will generally have no oversight by the developer in the event that other actors deploy the systems. Similarly, controls that are created can by bypassed via fine-tuning. This is perfectly reasonable, but the decision to release such systems despite these constraints requires arguments that either the systems cannot be misused in certain ways, or that their misuse by others isn't the responsibility of those developing and releasing the systems, or - as is usually the case in open source software - that the foreseeable misuse is small compared to the benefits of having open-source development. Of course, this may change in the near future for frontier models, as \citet{Meta2024FrontierAI} notes.

AI can also be deployed in places where control and oversight are unnecessary. For example, a game like AI Dungeon \cite{AIDungeon} is (at least arguably) a place where oversight or control is unneeded; the models used in the game have their own systems to prevent the creation of inappropriate or otherwise dangerous materials, and the usage within the game has little marginal risk. Many other applications of AI could allow developers to argue that the risks are minor or that the costs are excessive compared to the mitigated risk, which is certainly sometimes true. On the other hand, it is easy for developers to use the excuse that applications like entertainment are harmless without reflecting on the actual implications of the tools they produce. Therefore, a degree of caution is certainly warranted in claiming that any failures are unimportant or not cost-effective.

\section{Constructing Meaningful Oversight and Control of AI Systems}
Building on the above definitions and limitations of the different methods, we now endeavor to provide a clear and actionable framework for model builders, firms constructing or deploying AI systems, and governance decision-makers attempting to ensure that work was done properly. The framework may also be useful for auditors, regulators, and safety advocates to understand whether and how AI systems that have been deployed are meaningfully overseen.

As should be clear from the discussion, it is not always the case that supervision is possible. To ensure sufficient control and/or meaningful oversight of a system, a case based on the details of the systems and the risks must be made. That requires integration of the supervision design into a larger design process, as well as implementing the supervision itself in practice.

\subsection{The Design Process and Oversight}
We first very briefly review the basic system design process, as generally understood, for example, in \citet{dennis2012systems}, or specifically for ML systems, in \citet{huyen2022designing}. The process begins with planning and the analysis of requirements and use cases, then goes to design then implementation - which in the case of ML and AI systems is often done via iterative data collection and training, deployment, and finally operation monitoring, often along with continual learning and changes. Arguably, none of these stages are directly concerned with oversight and control, but ideally, all of them are. For this reason, frameworks explicitly targeted at risk management along the lifecycle have been proposed, especially for frontier AI systems. \cite{campos2025frontierairiskmanagement}

At the planning stage, risk analysis and identification should occur - though it is instead often (poorly) tacked on later. In parallel or following this, at the requirement stage, the requirements for a system must include how it is used and managed, including any planned oversight. For example, a critical use-case for any system being designed should be the oversight and governance of that system. This is the stage at which the oversight systems and methods are first proposed and designed. Additional practices at this stage might involve pre-mortems and the specification of metrics for assessment of whether the supervision is effective, and explicitly chosen checkpoints for when and how the systems will be reviewed.

In the case of frontier AI development, some have argued that this should be in place before training begins. \cite{campos2025frontierairiskmanagement} A draft of the EU AI Act Code of Practice even mandates "assessing and potentially mitigating
systemic risks" happening "potentially during the design and prototyping phases, but at the latest before starting a final training run." \cite{EC2024GPAICode} This means that the oversight design may need to be in place before other parts of the iteratively-developed system design.

The overall design should not only address the planned usage, but also reflect the planned ability to manage and supervise the system, for instance, ensuring that tools for oversight are compatible with the system as it is constructed. During iterated development, that developers of at least AI models with systemic risks are expected to perform "systemic risk assessment and mitigation along the entire model lifecycle," \cite{EC2025GPAICode} and specifically during training, "assessing and potentially mitigating systemic risks at pre-defined milestones." \cite{EC2024GPAICode} While it is not mandated, other ML and AI systems developers should at least understand and plan to mitigate risks at this stage.
Next, deployment starts the need for actual implementation of the planned supervision and management. Even before launch, however, some have suggested pre-deployment disclosures \cite{sherman2023airiskprofilesstandards} and the release of model cards no later than when the model is released. \cite{mitchell2019model} These would both presumably include details of the oversight process. 
Finally, during and following the internal or external deployment of an AI model or system, the oversight methods would be in full effect. This stage might seem to be purely operational, but it is also the critical time for assessment of the system in practice. Some parts of oversight build this in explicitly, especially passive oversight used for governance review, but systems reliant on other methods should be careful to ensure that oversight of the oversight or control system itself occurs. Evaluation of the types of errors which the oversight system anticipated,  and those that it did not, should occur, including a review of how the issues are handled and whether the planned control is working well. Additionally, routine re-evaluation of the system and the goals is critical; if checkpoints for routine evaluation of the supervisory systems were not put in place during the planning stage, they should certainly be specified now.

Of course, as with any production system, new needs, uses, or changed environment will necessitate adjustments in the system. It is easy to neglect the oversight at this stage, assuming that the extant process will be sufficient; obviously, however, this may not be the case. Special care is therefore needed after deployment, and during any changes in the use of the system by either the deployer, or by the users, to ensure that the system does not change in ways that render control or oversight no longer effective. For example, oversight of a self-driving car that assumes a possible driver is present who can take over if the system is rendered unsafe, say, by rain that reduces system visibility, or a sensor failure, could be invalidated if the system is used for transporting children. A system that has a failsafe behavior of pulling to the side and awaiting instructions assumes some observer will be present to note if an error occurs, which could no longer be true if the service is used for deliveries without a person inside. Therefore, if and when such a system's usage changes, the operators of the system need to re-assess the way in which oversight operates.

Thus far, we have suggested that the system must be designed, but have not provided an explanation of what specifying that design requires. To do so, we first outline the relevant dimensions and note the characteristics that need to be specified when documenting whether and how a system is supervised using each method. Obviously, in most systems some overlapping combination of methods for overlapping sets of risks will be appropriate, and the conceptual map we provide should help bolster understanding of where each method is intended to apply.

\subsection{Relevant Dimensions and Prerequisites}
As noted above, \citet{VerdiesenEtAl2021} lay out the time dimension, which we will label ex-ante, in-flight, and ex-post, and the layer dimension, technical, sociotechnical, and governance. Next, we note the different types of direct human involvement, namely none, Human in the Loop, Human-on-the-loop, and human-after-the-loop. Each of these three dimensions applies to the supervision method. 

However, supervision is not an independent function, and exists to address a specific risk or group of risks, which must be specified as well. As is standard, risk mitigation, which supervision assists, is only one step in a broad process for identifying the risks and defining and implementing the mitigation strategy. We therefore depend on a broader risk management process. Inputs into the process here include the risk identification, risk analysis, and identification of threat models or failure modes. \cite{NIST2023, campos2025frontierairiskmanagement}  While the full process is critical as an input to documenting oversight of the risks, it is also beyond the scope of the current discussion. We will, however, note that the risks which are dealt with in AI and that require oversight will span safety, ethics, and security, and each of these can and should be within scope for risk identification, and therefore for supervision. \cite{qi2024airiskmanagementincorporate}

Additional dimensions are a product of the above. That is, the combination of the above characteristics, namely the risk addressed, time-scope of the mitigation, layer of oversight, and method or degree of human involvement, will determine the final set of characteristics, discussed earlier in the paper. That is, the speed of failure, the severity of harms, and the detectability of the failures would therefore be specified for each combination of risk and oversight method.

Until now, we have not dealt with several critical practical concerns. First, the discussion primarily addresses the developers - but they are not the only stakeholder. Developers alone cannot define what qualifies as sufficient or meaningful oversight, and as we discuss below, the risk management process requires transparency to external oversight, which needs at minimum, external transparency and review. Ideally, it also extends to ensuring external input into the classes of risks mitigated, though again, this is primarily the responsibility of the broader risk framework. 

\subsection{Documenting the Supervision Method(s)}
Many proposed frameworks for risk documentation exist that can be adapted to oversight and control. We do not review or discuss these, but provide a structured schema of how to report the classes of oversight and control used as mitigations, which can (and should) be used as a basis for documentation.

\rule{0.4\textwidth}{0.4pt} \\
\emph{\textbf{Documentation Schema}}
\label{descr:documentation} 

\begin{description}[style=nextline,itemsep=1pt,parsep=0pt,leftmargin=0.6cm,labelsep=0.6em]
  \item[Oversight / Control] Name or brief description of the method used.
  \item[Time-scope and Type] Ex-ante, In-flight, or Ex-post, and type of control (e.g., “Control – Technical”).
  \item[Purpose / Risks Mitigated] Risks from the risk register which are addressed.
  \item[Human Involvement] Description of human involvement.
  \item[Feasibility Conditions] Assumptions or system characteristics that make the method viable.
  \item[Failure Modes] How the method can fail or risks may be unaddressed.
  \item[Review Plan] How and when the method is reviewed.
\end{description}
\rule{0.4\textwidth}{0.4pt}

To assist in filling out the schema, the two tables below review the classes of control and oversight strategies, respectively, along the identified dimensions. The notes additionally provide examples, necessary enablers for feasibility, and example failure modes. This is not intended to be complete, but should be indicative. That is, individuals documenting a specific system can use this as a starting point to reflect on the details of the system. Of course, as the template should make clear, it is still necessary to ensure that any additional relevant failure modes are identified, additional relevant constraints or feasibility concerns can be raised, and so on.

\section{Evaluating Supervision}

The ways in which oversight fails are critical, and deserve some additional attention. We discuss them briefly, then outline a maturity model for the types of control and oversight outlined in the paper.

The first and most critical point is that meaningful supervision cannot always occur; supervision may not suffice to address the risks. Therefore, a system developer or deployer must make a case that the oversight and control methods put in place are sufficient given the risk. That is, simply stating there is a human in the loop does not demonstrate that the system is under meaningful supervision. Among other things, the argument claiming that the method used is sufficient needs to explain the risks being addressed and how the classes of failure outlined above are handled. This may be part of a "safety case" as advocated by UK AISI and others.\cite{aisi2024safetycases} Alternatively, it may be a more informal or narrower explanation specific to the risk included in the model card. In either case, this must include clear identification of the risk and the specific supervision methods, such as in the format outlined above.

As a result of the need to make such an argument, the claims of oversight or control of specific risks would include public or externally reviewed documentation \cite{homewood2025thirdpartycompliancereviewsfrontier} and a clear explanation of how the systems put in place are sufficient. Otherwise, the claims are (unfortunately) indistinguishable from performative safety-washing, rather than trustworthy substantive risk mitigation.

\begin{landscape}

\noindent
\begin{minipage}{\linewidth}
  \captionsetup{justification=raggedright,singlelinecheck=false}
  \captionof{table}{Control Mechanisms}
  \label{tab:control-mechanisms}
  \renewcommand{\arraystretch}{1.1}
  \begin{tabularx}{\linewidth}{@{}L{2.5cm} L{2cm} L{5.5cm} L{6cm} L{5cm}@{}}
    \toprule
    Method & Time-Scope & Examples & Enablers & Failure Modes \\
    \midrule
    Technical \nobreak Control & Ex-ante
      & Programmed constraints; sandboxes; rule‑based limits; control systems; formal verification
      & Fully specified environment \& provable world‑model; sufficient observability
      & Specification gaps; control system failures; adversarial bypass \\
    \addlinespace[0.6ex]
    Human Control & Ex-ante, In-flight
      & Human-in-the-loop; variable‑autonomy protocols; shutdown/lock‑out procedures for worst-case risks (e.g., \citet{mickens2025guillotinehypervisorsisolatingmalicious})
      & Human attention; aligned incentives; training; interpretable/explainable AI; adequate human‑machine supervision interfaces; ability to identify loss‑of‑control
      & Alert fatigue; failures of attention/expertise/situational awareness; skill atrophy \& routinization; boredom; slow failure detection \\
    \bottomrule
  \end{tabularx}
\end{minipage}

\vspace{3.2em}

\noindent
\begin{minipage}{\linewidth}
  \captionsetup{justification=raggedright,singlelinecheck=false}
  \captionof{table}{Oversight Mechanisms}
  \label{tab:oversight-mechanisms}
  \renewcommand{\arraystretch}{1.1}
  \begin{tabularx}{\linewidth}{@{}L{2.5cm} L{2cm} L{4cm} L{7cm} L{5cm}@{}}
    \toprule
    Method Type & Time-Scope & Examples & Enablers & Failure Modes \\
    \midrule
    Preventive Active & Ex-ante
      & Red-teaming; predeployment audits; internal/external review
      & Management \& cross-stakeholder buy-in; technical expertise; time before deployment; precise risk definitions
      & Failure to anticipate failures; treadmill effect as models improve; focus on easily tested failure modes \\
    \addlinespace[0.6ex]
    Human-above-the-Loop & In-flight
      & Tiered escalation; adaptive autonomy
      & Dynamic policy; continuous monitoring; clear failure definitions
      & Institutional inertia; slow response; inability to identify or rectify failures \\
    \addlinespace[0.6ex]
    Management Review & Ex-post
      & Logging \& anomaly detection; postmortems; management reporting; public disclosures
      & Reliable instrumentation; assurance/audit resources; ombuds/whistle-blower protection; strong safety culture; independent reporting \& escalation channels
      & Detection latency; irreversibility; internal pressure; lack of transparency; liability concerns; incorrect root-cause diagnosis \\
    \addlinespace[0.6ex]
    Internal Governance & Ex-post
      & Review/audit systems; access restrictions
      & Robust safety culture; external accountability; enforceability
      & Jurisdictional gaps; lack of enforcement; routinization \\
    \addlinespace[0.6ex]
    External / Regulatory Oversight & Ex-post
      & Liability regimes; government/international oversight; third-party compliance reviews
      & Established in advance; early external knowledge of risks \& failures
      & Detection latency; irreversibility \\
    \bottomrule
  \end{tabularx}
\end{minipage}

\end{landscape}

Second, metrics are important for management in general, and oversight in particular, but developing useful such metrics pose particular challenges \cite{Manheim2023building}. The development of metrics for GenAI models themselves is still a rapidly evolving field \cite{weidinger2025evaluationsciencegenerativeai}, and the development of metrics for meaningful oversight is in many ways even more rudimentary. This is critically different from measurement of the risks themselves, but likely requires progress on that front as well. That said, even simple metrics, like Coverage $\times$ Detection $\times$ Remediation rates, can be useful in situations where no other quantitative process for measuring or evaluation of supervision exists.

Next, as a complement to metrics, continuous review of the process is critical, especially because the process can fail for a variety of reasons which must be understood and addressed. Identifying failure modes and addressing them is not a one-time process, though some of the failure modes should be known in advance, as were outlined. 
Despite the repetition, to mitigate the critical risk of not emphasizing the failure modes sufficiently, we review a few of the most critical general concerns. 

For control and some oversight methods, we can have issues with speed mismatches, when time‑to‑harm is less than the control system reaction time, or in the case of human involvement, less than the time needed to understand and react. There are also problems with under-specified or ambiguous contexts that can include novel scenarios outside training or procedure scope. For oversight, failure modes range from limited human capacity, including attention, expertise, situational awareness, to organizational incentives such as against raising red flags.

Notably, the organizational failures for some oversight methods need not be within the organization, such as reliance on oversight bodies influenced by industry or relying on external laws which do not keep pace with capabilities. Similarly, technical oversight issues can include a failure of technical transparency, such as insufficient interpretability and missing or limited audit methods or tools, insufficient logging or log analysis. 

Lastly, we note that frontier AI systems have the potential for creating "unknown unknown" risks, which challenges the entire premise that ex-ante risk identification is possible, which gives a reason to question whether oversight and control strategies broadly are sufficient. At the very least, horizon scanning and broader awareness of risks identified at other firms is needed. 

\subsection{Maturity Model}
In practice, the idealized process we present, which outlines the mitigations and risks addressed clearly and then implements an effective oversight process, may not be achieved. While successful prevention of a harm is binary, strategies and management approaches to mitigate the risks are not. For that reason, we suggest a maturity model for AI supervision. This should be useful, for example, for system designers making claims about their mitigations, and for auditors or regulators assessing the systems.

\begin{table}[h]
\caption{AI Supervision Maturity Model Summary} 
\label{tab:maturity_model}
\centering 
\begin{tabular}{@{} >{\raggedright\arraybackslash}p{0.05\textwidth} >{\raggedright\arraybackslash}p{0.4\textwidth} @{}} 
\toprule
\textbf{Level} & \textbf{Description} \\
\midrule 
 1 & \small Teams do not define risks or use control or oversight methods without measurement or clear view of how risks are addressed. No public documentation of their control process is provided. Any controls put in place should be considered safety-washing, to provide false assurance which may inflict additional harms.\\
2 & \small Teams pick and publicize mitigation tactics based on a patchy view of risks they may have a partial or complete risk register, but use supervision methods that do not address key risks. This often includes incorrect claims that risks are addressed by oversight, or suggestions that risks are not yet critical enough to require safety measures. \\
\small 3 & \small Teams map risks and supervision methods publicly and correctly, drawing from an appropriate variety of sources for risks to address, but don't have a full oversight or control process that considers and monitors all important risks. \\
\small 4 & \small Teams pair risk identification and trade-off analysis with supervision and appropriate mitigations, which are documented publicly along with the model, but only in some development lifecycle phases, and/or without continuous monitoring and publicly disclosed, well defined, and prespecified review and re-assessment systems.\\
\small 5 & \small Teams implement supervisory or control mitigations for all appropriate risks identified internally, as well as those disclosed in similar systems. Along with clear publicly available risk identification, there is public documentation of mitigations before even internal deployment, including monitoring and risk prioritization. This also acknowledges unknowns and monitors for new uses, risks, and system changes, and prepares to adapt to residual risk.
\vspace{2mm}
\end{tabular}
\begin{description} \small
 \item The five numbered levels here correspond to the labeled list of latent, emerging, developing, realizing, and leading in the Microsoft Responsible AI Maturity Model, which are less descriptively relevant here.
\end{description}
\rule{\linewidth}{0.4pt} 
\end{table}

In Table \ref{tab:maturity_model}, we suggest an adaptation of one dimension of the Microsoft Responsible AI Maturity Model, that for mitigating AI risks. \cite{vorvoreanu2023responsible} This can be used alongside a broader review of overall maturity, but can also be used as an independent evaluative tool.

\section{Conclusion}
In addition to clarifying terminology in ways that support standards processes and regulation, we have provided a clear path forward for better documentation and understanding. This is important for those developing, deploying, testing, or evaluating the purchase of models. However, we caution that not all risks from AI can be resolved with oversight or control. In addition, ethical dilemmas posed by AI are not resolvable through better oversight mechanisms alone. They also cannot address whether certain AI applications should be developed or deployed at all, which requires broader risk management, and societal discussions. 

The requirements for different strategies highlight that such oversight is not easy, especially if used in real-time and at scale, and when the system itself is poorly understood. Therefore, meaningfully controlling or supervising a given system might not be possible. Still, oversight and control play a useful role, and the above discussion enables better discussion, design, and operational management of supervision systems.

For this reason, our conclusion is hopeful for the better management of limited and well-understood systems, and to address the classes of risks which can be addressed. Outside of those systems, we reiterate that we advise caution against the idea that either control or oversight is a sufficient solution for ensuring their safety, or even a viable risk-mitigation strategy for many proposed future autonomous systems, especially those which pose unknown-unknown risks. \\

\bibliography{references} 

\end{document}